\documentclass[10pt,journal,compsoc]{IEEEtran}
\usepackage{mathrsfs}
\usepackage{algorithm}
\usepackage{algpseudocode}
\usepackage{bm}
\usepackage[font=small,labelfont=bf,tableposition=top]{caption}
\usepackage{graphicx}
\usepackage{subfigure}
\usepackage{amssymb}
\usepackage{amsmath}
\usepackage[table]{xcolor}
\usepackage{array,color}
\usepackage{arydshln}
\usepackage{verbatim}
\usepackage{booktabs}
\usepackage{tikz}
\usepackage{ctable}
\usepackage{multirow}
\usepackage{multicol}
\usepackage{rotating}
\usepackage{threeparttable}
\usepackage{cite}
\usepackage{pbox}
\usepackage{soul}
\usepackage[colorlinks=true,linkcolor=red,citecolor=blue]{hyperref}
\usepackage{url}
\usepackage{float}

\usepackage{multirow}
\usepackage{paralist}%
\usepackage{enumitem}
\usepackage{xspace}
\newcommand{\mbf}{\mathbf}

\newcommand{\degree}{\ensuremath{{^\circ}}}

\newcommand{\Baseline}{{\em Baseline}\xspace}
\newcommand{\LinAd}{{\em Lin-Ad}\xspace}
\newcommand{\SVRAd}{{\em SVR-Ad}\xspace}
\newcommand{\DiffSiam}{{\em Diff-NN}\xspace}

\newcommand{\DiffSiamout}{{\em Diff-NN$^{wo}$}\xspace}
\newcommand{\DiffVGG}{{\em Diff-VGG}\xspace}
\newcommand{\DiffFT}{{\em Diff-NN-Ad}\xspace}

\newcommand{\archi}[1]{{Sys-{#1}}\xspace}

\newcommand{\eyediap}{{EYEDIAP}\xspace}

\newcommand{\TrainingData}{\ensuremath{{\cal D}}\xspace}
\newcommand{\TestData}{\ensuremath{{\cal D}_{Test}}\xspace}
\newcommand{\RefData}{\ensuremath{{\cal D}_c}\xspace}
\newcommand{\NRefData}{\ensuremath{N_c}\xspace}
\newcommand{\GazeGT}{\ensuremath{\mbf g^{gt}}\xspace}
\newcommand{\GazePred}{\ensuremath{\mbf g^{p}}\xspace}
\newcommand{\GazeAdap}{\ensuremath{\mbf g^{ad}}\xspace}
\newcommand{\GazeSiam}{\ensuremath{\mbf g^{sm}}\xspace}

\newcommand{\DifPred}{\ensuremath{\mbf d^{p}}\xspace}

\newcommand{\headps}{\ensuremath{\mbf h}\xspace}

\newcommand{\SubjectIndex}{\ensuremath{k}\xspace}

\newcommand{\Image}{\ensuremath{I}\xspace}
\newcommand{\Jmage}{\ensuremath{J}\xspace}
\newcommand{\Fmage}{\ensuremath{F}\xspace}

\newcommand{\mypartitle}[1]{\vspace*{1mm}{\noindent {\bf #1}}}

\begin{document}

\title{A Differential Approach for Gaze Estimation}

\author{Gang Liu,
		\and Yu Yu,   
        \and Kenneth A. Funes Mora, 
        \and Jean-Marc Odobez*\\[-5mm]

\thanks{Gang Liu, Yu Yu, Jean-Marc Odobez are with the Idiap Research Institute, CH-1920, Martigny, Switzerland. E-mail: \{gang.liu, yyu, odobez\}@idiap.ch.}
\thanks{Jean-Marc Odobez is the corresponding author.}
\thanks{Kenneth Alberto Funes Mora is with Eyeware Tech SA, CH-1920, Martigny, Switzerland. E-mail: kenneth@eyeware.ch.}
}% <-this 
\markboth{PAMI short}%
{Shell \MakeLowercase{\textit{et al.}}: Bare Demo of IEEEtran.cls for Computer Society Journals}

\IEEEtitleabstractindextext{%

\begin{abstract}
Most non-invasive gaze estimation methods regress gaze directions directly from a single face or eye image.
However, due to important variabilities in eye shapes and inner eye structures amongst individuals,
%  universal models  obtain limited accuracies and their output usually exhibit high variance as well as biases which are subject dependent.
universal models  obtain limited accuracies and their output usually exhibit high variance as well as subject dependent biases.
  Thus, increasing accuracy is usually done through calibration, allowing gaze predictions for a subject 
  to be mapped to her actual gaze.
  In this paper, we introduce a novel approach, %image differential method for gaze estimation.
  % We propose to
  which works by   directly training a differential convolutional neural network 
  to predict gaze differences between two eye input images of the same subject.
  Then, given a set of subject specific calibration images, we can use the inferred differences to predict the gaze direction of a novel eye sample. 
  The assumption is that by comparing  eye images of the same user, annoyance factors (alignment, eyelid closing,
  illumination perturbations) which usually plague single image prediction methods can be much reduced, allowing better
  prediction altogether.
  Furthermore, the differential network itself can be adapted via finetuning
  to make predictions consistent with the available user reference pairs.
  Experiments on 3 public datasets validate our approach which constantly
  outperforms state-of-the-art methods even when using only one calibration sample or those relying on
  subject specific gaze adaptation.
\end{abstract}

\vspace*{-10pt}
}

\maketitle

\IEEEdisplaynontitleabstractindextext

% \fi
\vspace*{-50pt}
\IEEEpeerreviewmaketitle

\vspace*{-1mm}

\section{Introduction}
\label{sec:intro}
% As a non-verbal behavior and major indicator of human attention,
Gaze is an important cue of human behaviours. Gaze directions and gaze changing behaviours (such as gaze aversion, the intentional redirection away from the face of interlocutor~\cite{Andrist:2014:CGA:2559636.2559666}) are good indicators of the visual attention
and are also related to internal thoughts or mental states of people. 
Besides, as a non-verbal behaviour, gaze is an important communication cue which has also
been shown to be related to higher-level characteristics such as personality.
It thus finds applications in many domains like
Human-Robot-Interaction (HRI)\cite{Andrist:2014:CGA:2559636.2559666,Moon:2014:MMI:2559636.2559656},
Virtual Reality~\cite{Pfeiffer2007TowardsGI}, social interaction analysis~\cite{Ishii:2016:PNS:2896319.2757284},
or health care~\cite{vidal12_comcom},
%
%With the development of sensing function on mobile phones, gaze is also expected to
%be involved in a wider set of application in
or mobile phone scenarios~\cite{Krafka2016, tonsen17_imwut, huang2015tabletgaze}.

\mypartitle{Motivation.}
Non-invasive vision based gaze estimation has been addressed  with two main paradigms:
geometric models  and  appearance~\cite{hansen2010eye}.
%
%% The former ones rely on eye features such as eyelids or iris center
%% to learn a geometric model of the eye and  infer gaze~\cite{Wood2014,Alberto2014,valenti2012combining,sun2014real}.
Since the former
%ones rely on eye features such as eyelids or iris center
%to learn a geometric model of the eye and  infer gaze~\cite{Wood2014,Alberto2014,valenti2012combining,sun2014real}.
%
%But feature extraction
suffers from noise, image resolution, illumination, or head pose issues, 
appearance-based methods which predict gaze directly from the eye (or face) images have attracted more attentions in recent
years~\cite{zhang2015appearance, Sugano2014, Zhu_2017_ICCV, FunesMora2016}.
Among them, deep neural networks (DNN) have been shown to work well.

Nevertheless, even when using DNN regressors,  their 
accuracy
%these  method
has been limited to around $5$ to $6$ degrees, with a high inter person
variance~\cite{zhang2015appearance, Sugano2014, Zhu_2017_ICCV, FunesMora2016,shrivastava2017learning,Zhang2017, zhang2017s}.
This is due to many factors including dependencies on head poses, large eye shape variabilities,
and only very subtle eye appearance changes when looking at targets separated by such small angle differences.

\begin{figure}
	\vspace{-40pt}
  \centerline{
a)\includegraphics[width=0.45\linewidth]{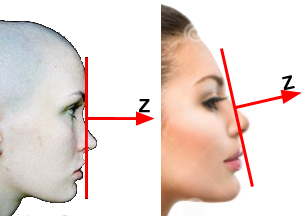}
b)\includegraphics[width=0.46\linewidth]{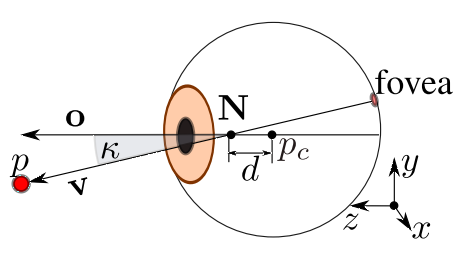}
  }
  \vspace{-3px}
  \caption{
    Examples of variability factors. 
    (a) Head pose shape variabilities induce different frontal head pose definition and hence variabilities in eye images.
     (images  from Pinterest.com).
    (b) Variabilities across subjects of the difference  between the visual axis (unobserved, defining gaze)
    and the optical axis (defined by iris center, observed) introduces gaze prediction uncertainties (image from~\cite{FunesMora2014}). 
  }
  \vspace{-12px}
  \label{fig:prob_b}
  \label{fig:prob_axis}
  \label{fig:prob_c}
\end{figure}

\begin{figure}
  \centerline{
    \includegraphics[width=0.8\linewidth]{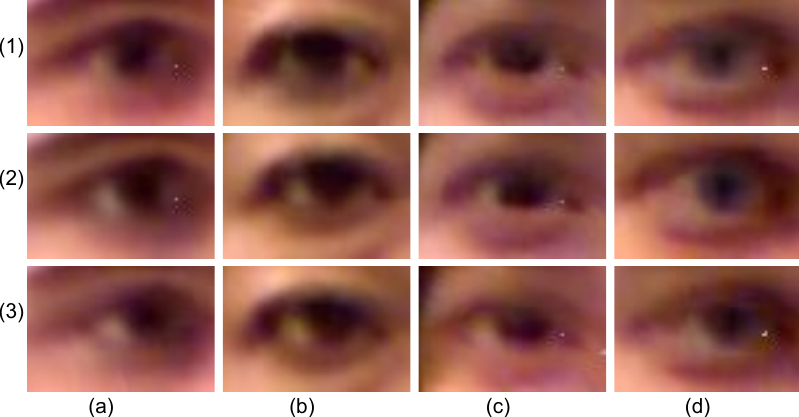}
    }
	\vspace{-5px}
	\caption{Appearance comparison.
          Columns (a) to (d) show right eye images from different persons from the \eyediap dataset~\cite{funes2014eyediap}.
          Row (1) to (3) correspond to gaze directions with the same pitch (5 degrees) and a yaw of 5, 10, 15 degrees respectively.
        }
	\label{fig:prob}
	\vspace{-10px}
\end{figure}

For instance, Fig.~\ref{fig:prob_b}(a) shows the difficulty to define
an absolute head pose like a frontal pose. This has a non negligible
impact on the eye appearance.
Another factor explaining % why appearance based methods encounter
the limited accuracy when building person independent models
is that the visual axis is not aligned with the optical axis (related to the observed iris) \cite{guestrin2006general},
and that such alignment differences are subject specific (see Fig.~\ref{fig:prob_c}(b)), with a standard deviation
of 2 to 3 degrees amongst the population without eye problems.
%
%that the Kenneth et al.~\cite{FunesMora2014} reveals that the visual axis of gaze is the line connecting the visual object and the nodal point which is an internal point of eyeball and whose position is person specific.
Said differently, in theory, images of two eyes with the same appearance but with different internal eyeball structure can correspond
to different gaze directions, demonstrating that gaze can not be fully predicted from the visual appearance.
Altogether, in practice, such variabilities introduce confusions for regression, as illustrated in 
Fig.~\ref{fig:prob} which shows that gaze related elements (like iris location or the eyelid closing)
in eye images from different persons  sharing the same gaze directions
%(same pitch and yaws between 5 and 15 degrees)
can look quite different, while 
%
%The yaw direction in Row (1) to (3) are 5, 10, 15 degrees respectively.
more importantly,  eye of different persons can be similar when they
look at different directions (see  (a-2) and (d-3)). 

A straightforward solution to this problem is to learn person-specific models~\cite{sugano2014learning,zhang2015appearance,tHoser2016personalization} or fine-tune a pre-trained model~\cite{Masko2017}.
%which can achieve far better accuracy.
%
Note that even regular high-end Infra-Red (IR) devices (eg from Tobii)
require users to stare at several fixed positions before using them.
However, training person-specific appearance models may require large amounts of personal data,
especially for DNN methods and even when conducting simple network fine tuning adaptation.
%
%
%As this is not practical in real life applications,
Other methods rely on fewer reference samples to train a linear regression model~\cite{Lu2014} or an SVR~\cite{Krafka2016}.
Still these methods are usually not robust to environment changes and their accuracy is  heavily affected by the number of reference samples. 
A too small amount can even result in worse performance than without calibration. 
%If it is not large enough, the calibrated results might be even worse than those without calibration.
This is unfortunate, as there are plenty of real scenarios in which we can collect a few annotated samples.

%\begin{figure}
%\centering
%	\includegraphics[width=0.24\linewidth]{images/difgaze_fig/ref1.png}
%	\includegraphics[width=0.24\linewidth]{images/difgaze_fig/ref5.png}
%	\includegraphics[width=0.24\linewidth]{images/difgaze_fig/ref8.png}
%	\includegraphics[width=0.24\linewidth]{images/difgaze_fig/ref15.png}\\\vspace{2px}
%	\includegraphics[width=0.24\linewidth]{images/difgaze_fig/test1.png}
%	\includegraphics[width=0.24\linewidth]{images/difgaze_fig/test5.png}
%	\includegraphics[width=0.24\linewidth]{images/difgaze_fig/test8.png}
%	\includegraphics[width=0.24\linewidth]{images/difgaze_fig/test15.png}
%\caption{From left to right column: eye images from different person from the \eyediap dataset~\cite{funes2014eyediap}. Gaze groundtruth (yaw and pitch) on the top and bottom row are $[0,0]$ and $[10,0]$ respectively.}
%\label{fig:prob}
%\end{figure}

\begin{figure}
	\centering
	\includegraphics[width=1\linewidth]{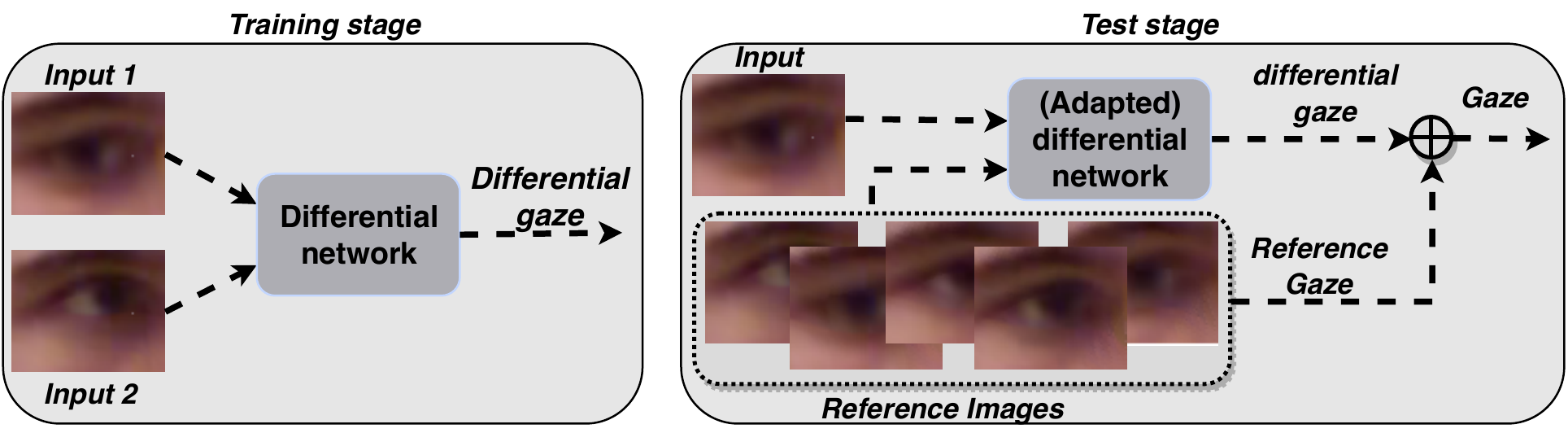}
	\vspace{-15px}
	\captionof{figure}{Approach overview.
		During training, random pairs of samples from the same eye are used to train a differential network.
		At test time, given a set of reference samples, gaze differences are computed
		and used to infer the gaze of the input image.
                To gain higher accuracy, the differential network can be adapted via fine-tuning using the pairs of reference samples.
	}
	\label{fig:siamese_framework}
	\vspace{-14px}
\end{figure}

\mypartitle{Contributions.}
This paper is an extension of our paper \cite{liu2018differential}.
we aim to solve the person-specific bias using a few annotated reference samples from the specific person.
To this end, two strategies have been considered and analyzed.
The first one is a baseline and 
%is  a simpler method than those mentioned  above for adaptation. 
%It
consists of learning the linear relationship between the gaze predictions from a pre-trained NN applied to few training samples and their
groundtruth gaze.
Interestingly, although simple, it is shown to achieve better results than the state-of-the art SVR method of~\cite{Krafka2016}.

%Secondly, although the previous methods can reduce the subject specific bias between the subject (test) data and the overall training dataset,
The second method corresponds to our main contribution, and is as follows.
Although the previous methods can reduce the subject specific bias between the subject (test) data and the overall training dataset,
it does this by only working with the gaze prediction or feature outputs,
and does not account for %the bias between identities within training set and
the high gaze prediction variance within each subject's data.
%
%To address this issue, our main contribution is to propose a differential gaze estimation approach,
To address this issue, we propose a differential gaze estimation approach,
by training a differential NN to predict the gaze difference between two eye images instead of predicting
the gaze directly.
We hypothesize such a differential approach is less problematic than predicting gaze because the person dependent error
(such as shape, alignment errors) will be alleviated.
This is illustrated in  Fig.~\ref{fig:prob}, which shows that given an eye of a person,
it is easier to judge whether it is looking more to the left or the right than a second eye image if the latter
come from the same person than if it comes from another person (even with a similar eye shape).
In Fig.~\ref{fig:prob}, it is easier to see that the eye images at the bottom look more to the right than
the eye images at top which are in the same column than compared to images in the other columns. 

Our framework is illustrated in Fig.~\ref{fig:siamese_framework}.
At training time, a unified and person independent differential gaze prediction model is built which can be used at test
time for person specific gaze inference relying on only a few calibration samples.

Thirdly, as there are many architectures that can be designed for differential gaze (early fusion by concatenating the two images, or fusion of feature maps of the two images at different levels), we investigate and compare different architectures and show that mid-level fusion is better than early fusion or late fusion. Usually they have different impacts on gaze estimation.
%In this paper, we compare the results using several different architectures.

\mypartitle{Paper organization.}
We discuss related works in Section~\ref{sec:relatework}.
In Section~\ref{sec:baisnn}, we introduce a state-of-the-art NN  for gaze prediction, 
illustrate the subject specific bias problem, and present  a baseline linear adaptation method to build
subject specific gaze prediction models.
In Section~\ref{sec:siamese}, we introduce our approach and the proposed modified siamese NN for differential gaze prediction.
Experiments are presented in Section~\ref{sec:exp}, while Section~\ref{sec:con} concludes the work.

\begin{figure}[tb]
	\centering
	\includegraphics[width=1\linewidth]{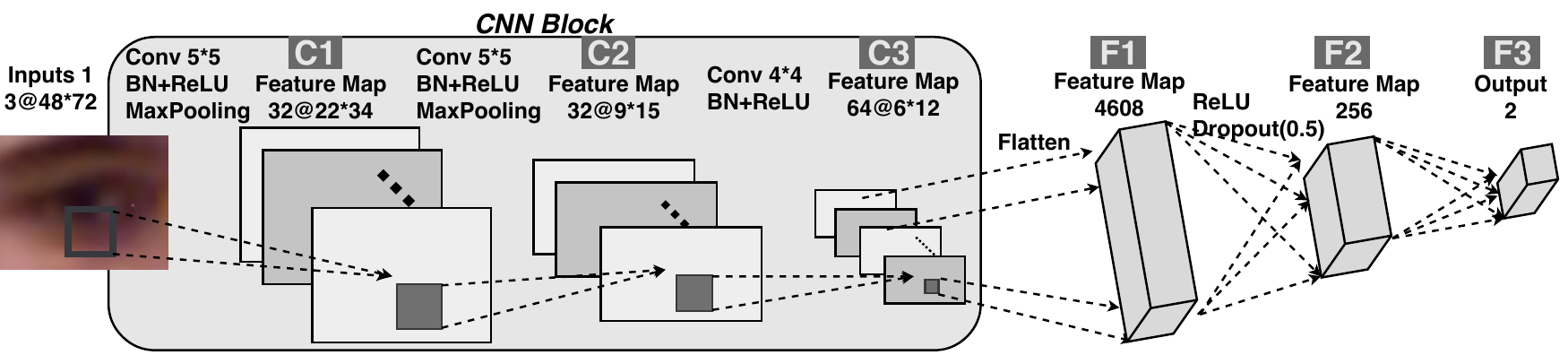}
	\vspace{-17px}
	\caption{Baseline CNN structure for gaze estimation. %Note the structure different from~\cite{Zhang2015}.
	}
	\label{fig:gazecnn}
	\vspace{-14px}
\end{figure}

\vspace*{-2mm}

\section{Related works}
\label{sec:relatework}

%\vspace*{-1mm}

Our work relates to   appearance-based modeling, person dependent calibration methods,
and to some extend, to siamese network approaches for achieving other tasks.

\vspace*{-2mm}

\subsection{Appearance-based gaze estimation}

%\vspace*{-1mm}

%Non-invasive vision based gaze estimation has been addressed using two main paradigms~\cite{hansen2010eye}: geometric models, and appearance. 
%
As said earlier, geometric approaches rely on eye feature extraction (like glints when working with infrared systems,
eye corners or iris center localization)
to learn a geometric model of the eye and then infer gaze direction using
these features~\cite{Wood2014,Alberto2014,valenti2012combining,sun2014real,Wood2016,Wang2017}.
However, they  usually require high resolution eye images for robust and accurate feature extraction,
are prone to noise or illumination perturbations,
and do not handle well head pose variabilities.
%and medium to large head poses.

Hence, many recent methods  rely on an appearance based paradigm ~\cite{zhang2015appearance, Sugano2014, Zhu_2017_ICCV, FunesMora2016},  
%directly predicting gaze from an eye (or face) image input
exhibiting more robustness with low to mid-resolution images
and obtaining good generalization performance.
%
%Amongst them, deep neural networks (NN) have been shown to work well. 
%
There, Neural networks (NN) methods have been shown to work well due to their ability to leverage  large
amount of data to train a regression network capturing the essential features of the eye images under various conditions
like illumination and self-shadow, glasses, impact of head pose.
For instance, \cite{zhang2015appearance} relied on a simple LeNet shallow network applied to eye images
and first demonstrated that NNs outperform most other methods.
%appearance based methods.
Very recently, a deeper pretrained network (VGG-16~\cite{Simonyan14c}) was fine-tuned for gaze estimation
and further improved the accuracy~\cite{Zhang2017a}.
%Due to this, many other CNN based methods have been proposed recently.
In other directions, Krafka  \textit{et. al}~\cite{Krafka2016} proposed to combine eye
and face  together using a multi-channel network, 
Zhang \textit{et. al}~\cite{Zhang2017} trained a weighted network to predict gaze from a full face image.
Shrivastava \textit{et. al}~\cite{shrivastava2017learning} learned a model from simulated eye images
using a generative adversarial network. 

\vspace*{-2mm}

\subsection{Person dependent calibration}
%
%As explained in the introduction,
Person dependent calibration is critical to obtain a more robust and accurate model for gaze estimation 
%
%(note that this is even the case for infrared head mounted device~\cite{lanata2015robust,mansouryar20163d,sugano2015self}).
%(this is alos the case for infrared head mounted device~\cite{lanata2015robust,mansouryar20163d,sugano2015self}).
(this is also the case for infrared head mounted device~\cite{lanata2015robust,sugano2015self}).
To solve this problem, Lu \textit{et.al}~\cite{Lu2014} proposed an adaptive linear regression method relying on few training samples,
but the eye representation (multi-grid normalized mean eye image) is not robust to environmental changes.
%
%% Starting from a trained NN, Krafka \textit{et.al.}~\cite{Krafka2016} relied
%% on eye images when looking at a grid of $13$ dot sample.
%% Feature maps from the last layer of the pretrained NN were then employed to train a Support-Vector-Regression (SVR)
%% person specific gaze prediction model.
%% However, SVR regression from a high dimensional feature vector input is not robust to noise.
%
Starting from a trained NN, Krafka \textit{et.al.}~\cite{Krafka2016} relied on 
%on eye images when looking at a grid of $13$ dot sample.
feature maps from the last layer of a pretrained NN to train a Support-Vector-Regression (SVR)
person specific gaze prediction model from 13 reference samples.
However, SVR regression from a high dimensional feature vector input is not robust to noise.
Different from~\cite{Krafka2016}, Masko tried to fine-tune the last layer of a pre-trained model for each subject,
but this  requires large amounts of data.
%, which is not practical in real life application.
%
In another direction, Zhang \textit{et.al.}~\cite{Zhang2018}
proposed to train person-specific gaze estimators from user interactions with multiple devices,
such as mobile phone, tablet, laptop, or smart TVs, but this does not correspond
to the majority of use cases.

However, none of the above works have proposed to calibrate gaze by estimating gaze difference from reference images,
which as we show in this paper is a much more  robust approach requiring less reference images.

\vspace*{-2mm}

\subsection{Siamese network}

%Our modelling is inspired by
They have first been  proposed in~\cite{bromley1994signature} for signature verification, and 
%using image matching.
%
with the deep learning revival,  for tasks like 
%feature extraction~\cite{zagoruyko2015learning, simo2015discriminative, kumar2016learning},
feature extraction~\cite{zagoruyko2015learning,  kumar2016learning},
%image matching and retrieval~\cite{wang2015sketch, luo2016efficient},
image matching and retrieval~\cite{wang2015sketch},
one-shot recognition~\cite{koch2015siamese},
%person re-identification~\cite{mclaughlin2016recurrent,varior2016gated}.
person re-identification~\cite{varior2016gated}.
%, or object tracking~\cite{bertinetto2016fully}.
%
They  consist of two parallel networks with shared weights,
%in which a pair of distinct images is used as input, one for  each network, 
a pair of  images as input (one per network), 
%parallel channel,
and the distance between their outputs %of each parallel network
is  the siamese network output.
%
%Implicitly, when dealing with discrete category problems, the goal is to learn
Often, for classification, the goal is to learn
%(usually using a hinge-loss function)
%a mapping from the image space to a new feature space
an embedding space,
%mapping from the image space to a  feature space
where samples from the same class are close and samples from different classes are far.
%
%In the regression case (our case), the loss function is usually defined by comparing the output distance with the groundtruth one.
In regression, the loss function compares the output distance with the groundtruth one.
Venturelli \textit{et.al.}~\cite{venturelli2017depth} use such an approach for head pose estimation.
%One of the closest work to ours is Venturelli \textit{et.al.}~\cite{venturelli2017depth}.
%
However,
%they are addressing a different task (head pose estimation),
%and
%inspired by the works on face identification, 
%their goal is more to
they use a multi-task approach in which both absolute poses and head pose differences are used as loss function.
At test time, the pose is still directly predicted from a single image.
Hence, while several layers of our differential networks are used to predict the gaze difference,
in their case the pose differences was only computed from the network pose prediction output.

The few-shot learning approach of~\cite{sung2018learning} is closer to our work.
Authors rely on a relation network trained to compare images.
As for us, its architecture
%is also a variant of Siamese networks similar to ours,
%the the sense that their network
consists of an embedding module that extracts featuremaps of images,
and a relation module using the concatenated featuremaps as input
to calculate a relation score.
Their method however addresses a quite different task (image classification vs gaze regression),
%and rely on
with a different loss function
%(based on the best matching score, compared to gaze difference prediction),
and they do not further adapt the network using the reference samples.

\vspace*{-1mm}

\section{Baseline CNN approach and linear adaptation}
\label{sec:baisnn}

\begin{figure*}[thb]
	\centering
	(a)
		\includegraphics[width=0.44\textwidth,height=32mm]{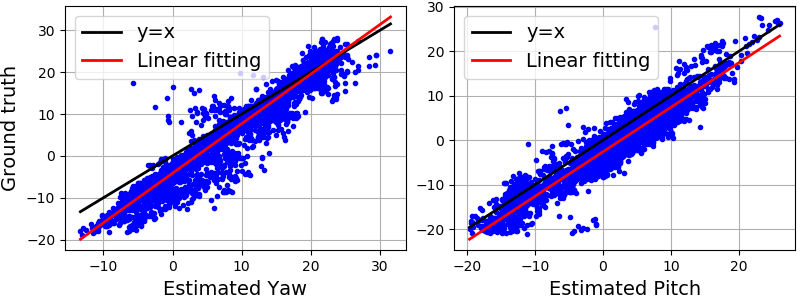}
	(b)
		\includegraphics[width=0.44\textwidth,height=32mm]{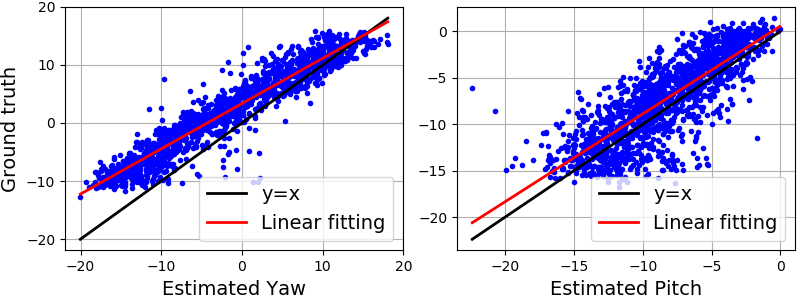} 
	\vspace{-4px}
	\caption{Scatter plot of the network regression (X-axis) and labelled groundtruth (Y-axis) of the yaw (left plot) and pitch (right plot) angles
		for an individual eye taken in the (a) \eyediap dataset; and (b) MPIIGaze dataset.
		%datafor network regression (X-axis) and labeled groundtruth (Y-axis). In each subfigure, the left and right are about yaw and pitch in gaze.
	}
	\vspace*{-15px}
	\label{fig:bias}
\end{figure*}

We first introduce a standard convolution neural network (CNN) for person independent gaze estimation.
We then show the resulting bias existing for unknown individuals,
and present a baseline linear adaptation method to solve it.

\vspace*{-2mm}

\subsection{Gaze estimation with CNN}

\vspace*{-1mm}

\mypartitle{Network structure.}
Fig.~\ref{fig:gazecnn} presents the standard NN structure for gaze estimation.
It consists of three convolutional layers and two fully connected layers\footnote{Note that it is slightly
  different from~\cite{zhang2015appearance}.}.
More precisely, the input eye image $\Image\in R^{M\times N\times C}$, where  $(M,N,C)=(48,72,3)$ denote
the dimensions and number of channels of the image, is first whitened. 
%standardization.
The convolutional layers are then applied and the resulting feature maps are flattened to be fed
into the fully-connected layers.
The predicted gaze direction $\GazePred(\Image) \in R^{2\times 1}$ is regressed at the last layer. 
The details of the network parameters can be found in the figure.

\mypartitle{Loss function.}
% For the identity independent task, we are given a dataset $\trset$ which consists of a series of identities as training set, and another dataset $\teset, \teset\cap \trset=\emptyset$ as test set.
%
% Suppose the labeled groundtruth of eye image $\imgai$ from identity $a$ is $\gai$, we can define a $\mathrm{L}1$ loss function as 
% \begin{align}
%     \mathcal{L} = \frac{1}{|\trset|}\sum_{i,p} \| \trdef{G}{p}{i} - \trtid{G}{p}{i} \|_1, \quad \trdef{I}{p}{i} \in \trset,
% \end{align}
% where $|\trset|$ denotes the number of elements in $\trset$.
Denoting the gaze groundtruth of an eye image \Image by $\GazeGT(I)$,
%and the gaze prediction model is given by $\GazePred(\Image)$,
we used the following $\mathrm{L}1$ loss function:
%to train our baseline NN:
\begin{equation}
    \mathcal{L} = \frac{1}{|\TrainingData|} \sum_{\Image \in \TrainingData} \| \GazePred(\Image)-\GazeGT(I) \|_1,
\end{equation}
where \TrainingData  denotes the training dataset and $|\cdot|$ denotes the cardinality operator.

\mypartitle{Network training.}
% \mynew{
% For eye images in the dataset, we first resize it into $48\times 72$ pixels. More precisely, we up-sample the image using bilinear interpolation if their size are smaller than this (MPIIGaze dataset). Otherwise, we randomly crop patch with this size around eyes (\eyediap dataset). The input in Fig.~\ref{fig:gazecnn} has three channels for color images, but we can adapt it into one channel for gray scale images. The eye images are first whitened before be sent to the network.
% }
For eye images in the dataset, we first resize them into a fixed resolution $s=48\times 72$. Concretely, we up-sample the images using bilinear interpolation if their sizes are smaller than $s$ (MPIIGaze dataset). Otherwise, we randomly crop patches with size $s$ around eyes (\eyediap dataset). The input has either three channels for color images as shown in Fig.~\ref{fig:gazecnn}, or one channel for gray scale images.
%The eye images are first whitened before being sent to the network.

%
The network is optimized with Adam method, with a learning rate initially set to $0.001$ and then divided by $2$ after each epoch.
In our experiment, $10$ epochs are applied and proved to be sufficient. The mini batch size is  $128$.

% \mypartitle{Network prediction.}
% Given a series of eye samples $\trdef{I}{q}{i}$ from identity $q$, where $q$ is the identity from test set $\teset$. For each sample $\trdef{I}{q}{i}$, we can predict its corresponding gaze $\trtid{G}{q}{i}$ using the neural network.

\vspace*{-3mm}

\subsection{Bias analysis and baseline linear adaptation method}

\vspace*{-1mm}

Because each individual eye has specific characteristics (including internal non-visible dimensions or structures), in practice,
we often observe a data bias between the network regression $\GazePred(\Image)$ and the labeled groundtruth $\GazeGT(\Image)$
of the eye images $\Image \in \TestData$ belonging to a single person.
This is illustrated in Fig.~\ref{fig:bias}, which provides a scatter plot of the $(\GazePred(\Image),\GazeGT(\Image))$ angle pairs in typical cases,
which can be compared with the identity mapping (black lines).

%Given the test set $\TestData$ that consists eye samples from a single identity, fig.~\ref{fig:bias} shows a scattering points using $\GazePred(\Image)$ as the X-axis and the $\GazeGT(\Image)$ as Y-axis.
%The left two are about gaze yaw and pitch respectively on one \eyediap identity. The right two are about gaze yaw and pitch respectively on one MPIIGaze identity.

%Usually the regressed gaze $\GazePred$ will be taken as the prediction. However there will be a large bias between the prediction and groundtruth according to our observation. Fig.~\ref{fig:bias} illustrates the bias between the black lines which indicate $y=x$ and the blue scattering points.

As can be observed, %from these typical data,
there is usually a linear relationship between $\GazeGT(\Image)$ and $\GazePred(\Image)$,
which is illustrated by the red lines in the plots.
Thus, when a set \RefData of sample calibration points of a user (usually 9 to 25 points) is available,
it is possible to learn this relation and obtain an adapted gaze model $\GazeAdap$ by fitting
%a simple solution will be doing a linear adaptation to reconstruct the relationship between $\GazePred(\Image)\in R^{2\times 1}$ and $\GazeGT(\Image)\in R^{2\times 1}$, with respect to the bias, by
a linear model
\begin{align}
\label{eq:linear}
    \GazeAdap(\Image) = A \, \GazePred(\Image) + B 
\end{align}
where $A\in R^{2\times 2}$ and $B\in R^{1\times 2}$ are the linear parameters of the model
which can be estimated through least mean square error (LMSE) optimization
using the calibration data. 

%\mynew{
%We are aware that gaze yaw and pitch are probably independent to each other, thus the parameter $A$ could be set as a diagonal matrix. But from the experiments we find out that associating yaw and pitch by using full matrix $A$ works better at most time, especially when they have less noise. We believe the reason under this circumstance is the robust of estimation on yaw and pitch can be improved mutually. 
%}
%% At first hand, the yaw and pitch can be considered as being independent of each other, and the parameter $A$ could be set as a diagonal matrix. However, in theory, user's eye might not have been normalized or rotated in some way, so that associating yaw and pitch by using full matrix should indeed be a better model. This is confirmed in practice especially when there is little noise.

\vspace*{-1mm}

\section{Proposed differential approach}
\label{sec:siamese}

\vspace{-1mm}

\subsection{Approach overview.}

\vspace{-1mm}

The linear adaptation above allows to correct biases from the gaze output, but this does not really account for the specificity of a user's eye, nor was the network trained to take into account the presence of biases. The method we propose aims at solving these issues.
It is illustrated in Fig.~\ref{fig:siamese_framework}.
Its main part is a differential network designed and trained to predict the differences in gaze direction
between two images of the same eye.
At test time, the gaze differences between the input eye image and a calibration set of reference images are computed first.
Then the gaze of the eye image is estimated by adding these gaze differences to the reference gazes.
The calibration set could further be used to adapt the  network
by fine-tuning so that is makes better differential predictions between reference sample pairs.
The details of the different components are introduced in the following paragraphs.

% \begin{figure}[!t]
%     \centering
%     \includegraphics[width=0.8\textwidth]{images/framework2.png}
%     \caption{Caption}
%     \label{fig:siamese}
% \end{figure}

\begin{figure}[tb]
	\centering
	\includegraphics[width=1\linewidth]{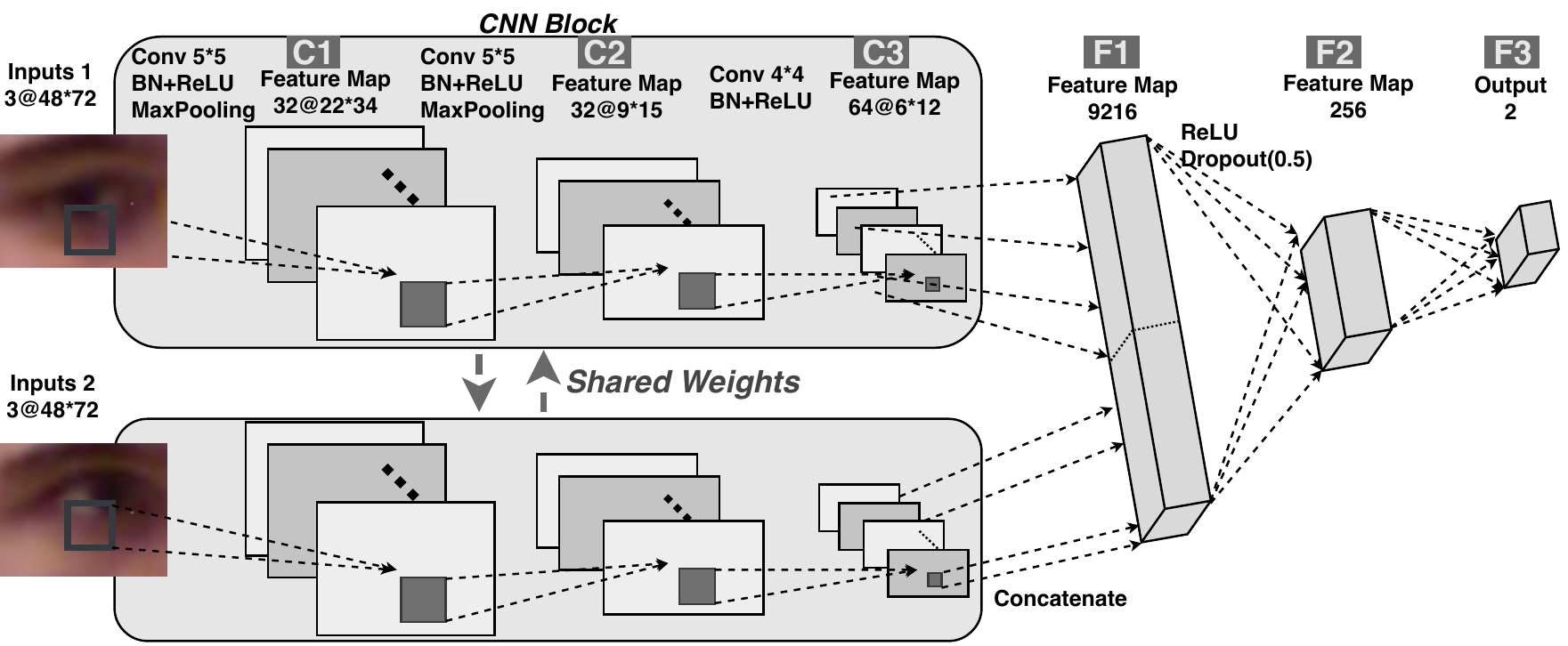}
	\vspace*{-14px}
	\caption{The designed differential network.}
	\label{fig:siamese}
		\vspace*{-12px}
\end{figure}

\vspace{-2mm}

\subsection{Differential network architecture.}
The network we use is illustrated in Fig.~\ref{fig:siamese}.
%, and is modified from the traditional siamese approach.
%
Each branch in the parallel structure is composed of three convolutional neural layers,
all of them  followed by batch normalization and ReLU units.
Max pooling is applied after the first and second  layers for reducing the image dimensions.
%
%The network consists of a parallel of convolutional layers with shared weights.
%The size of the kernels are precisely shown in the figures. All the convolutional layers 
%
After the third
%convolutional
layer, the feature maps of the two input images are flatted and concatenated into a new tensor.
Then two fully-connected layers are applied on the tensor to predict the gaze difference between the two input images.
Thus, where traditional siamese approaches would predict the gaze for each image,
and compute the differences from these predictions, our approach uses neural network layers to predict
this difference from an intermediate eye feature representation.

The architecture we propose has several advantages.
First, %given the image input
it is a good trade-off between prediction capacity and running-time.
Secondly, while we could directly provide the two images as input to the network, this could increase the
computational cost and not necessarily provide better prediction. We demonstrate this in the experimental section. 

\vspace{-2mm}

\subsection{Loss function, network training and adaptation}
\label{sec:lossadapt}

\vspace{-1mm}

The differential network is trained using a set of random image pairs $(\Image, \Jmage)$ coming from the same eye in the training data.
Denoting by  $\DifPred(\Image,\Jmage)$ the gaze difference predicted by the network,
we can define the loss function as:
\begin{align}
  \mathcal{L}_{diff} = \sum_{\Image, \,\Jmage \in \TrainingData^{\SubjectIndex}  }
   \| \DifPred(\Image, \Jmage) - \left( \GazeGT(\Image) - \GazeGT (\Jmage) \right)  \|_1,
	\label{eq:lossfunc}
\end{align}
where $\TrainingData^{\SubjectIndex}$ is the subset of $\TrainingData$ that only contains
images of the same eye of person \SubjectIndex.
% eye\footnote{Note: we learn separately a model for the left and the right eyes.} of person \SubjectIndex.

\mypartitle{Network training.}
Optimization is done with the Adam method and an initial learning rate of  $0.001$ which is 
divided by $2$ after each epoch.
In experiments, 20 epochs are applied. The mini batch size is  $128$.
To reduce the number of possible image pairs,
we have constructed the dataset of pairs by 
using each  image $\Image \in \TrainingData^{\SubjectIndex}$ as first image
and randomly selecting the second image $\Jmage$ in $\TrainingData^{\SubjectIndex}$.

%% Note that as the number of possible image pairs is too large,
%% we have reduced it  by using each  image $\Image \in \TrainingData^{\SubjectIndex}$ as first image
%% and randomly selecting the second image  $\Jmage \in \TrainingData^{\SubjectIndex}$ of the pair. 
%% So we have $|\TrainingData^{\SubjectIndex}|$ pairs for the subject \SubjectIndex.

\mypartitle{Network adaptation.}
At test time, since we are given a small calibration set $\RefData$ of reference images,
we can fine-tune our network by selecting  pair of samples $(\Image, \,\Jmage) \in \RefData$ and apply the same loss function ~\eqref{eq:lossfunc}.
In experiments, all possible pairs from \RefData were used, and the same fine-tuning of 10 epochs  with a fixed learning rate of 2e-4 was applied in all cases.
%Need to mention that the fine-tuning strategy is not necessary, but the icing on the cake.

\vspace*{-2mm}

\subsection{Gaze inference at test time.}

\vspace*{-1mm}

%
%As the network predicts gaze differences only, the method requires at least one reference image to predict an absolute gaze vector.
%
%In practice, we rely on a small calibration set $\RefData$ of images of the same eye.
%
Given the calibration dataset $\RefData$ of the user's eye,
we can first adapt the differential network (this is optional) as seen above.
Then, we use the network to  predict the gaze difference \DifPred(\Image, \Fmage) between the test image \Image and the reference images \Fmage,
and combine these gaze difference with the gaze groundtruth \GazeGT(\Fmage)
%of the reference images
to infer the gaze direction of the test image as \GazeGT(\Fmage)+ \DifPred(\Image, \Fmage). 
%
% Considering we do not estimate the gaze directly, we have to use some reference images for auxiliary. For the identity $q \in \TestData$, suppose we are given a set of reference images $\RefData$ from identity $q$. 
%
%Suppose $\TestData$ contains a single identity and $\RefData\subset \TestData$, for each eye sample $\Image\in \TestData$, we construct pairs of images by this sample and every reference image, which we denote as $(\Image, \Fmage), \Fmage\in \RefData$. By using pair $(\Image, \Fmage)$ as input of the network, we can predict the gaze difference $\DifPred(\Image, \Fmage)$ between image $\Image$ and each reference image $\Fmage$.
%
More formally:
\begin{align}
\label{eq:gazeest}
    \GazeSiam(\Image) = \frac{\sum_{\Fmage\in \RefData} w\left(\DifPred(\Image, \Fmage)\right)\cdot \left( \GazeGT(\Fmage)+ \DifPred(\Image, \Fmage)\right)}{\sum_{\Fmage\in \RefData} w\left(\DifPred(\Image, \Fmage)\right) },
\end{align}
where $w(\cdot)$ is  %the function
weighting the importance of each prediction.
%related to the gaze difference between test image \Image and the reference image \Fmage. 

Intuitively, if the reference eye image is more similar to the test eye image, we should be more confident about the gaze difference.
Thus the weight has been defined as a function of $\DifPred(\Image, \Fmage)$, which is a good indication of such similarity.
In practice, we simply use a zero-mean Gaussian  $\mathcal{N}(0,\sigma)$ as weight function.
%we assign a larger weight to the similar reference eye images. A simple way would be using a zero-mean Gaussian function $\mathcal{N}(0,\sigma)$ as the weight function. 
If $\sigma$ is too small, reference samples with large gaze difference will have no contribution. While if $\sigma$ is too large, reference samples will almost all have equal weights.
In experiments, $\sigma=0.1$ radian (5.7 degrees) has been used on all datasets,
although better values could be searched for per dataset using validation datasets within the training data.
%(see as well Section.~\ref{sec:weight}).

% We aware that the bias still exist on test state for unknown identity. Fig.~\ref{fig:bias2} shows the gaze prediction and the corresponding groundtruth for one identity in \eyediap dataset. We can see the minor bias...

% Fortunately we still can calibrate the results by using the reference image. If we apply the K reference images into Eq.~\ref{eq:gazeest}, we can obtain the estimated gaze for the reference images, which we denote as $\tilde{g}^{(a',phi(k))},k=\{1,\cdots, K\}$. 

% Apply Eq.~\ref{eq:adap_para} with $g^{(a',\phi(k))}$ and $\tilde{g}^{(a',k)},k=\{1,\cdots, K\}$ to estimate the parameters $\Tilde{A}, \Tilde{B}$ for adaptation. Thus, for the given test image sample $I^{(a',i)}$, we calculate the calibrated gaze $\hat{g}^{(a',i)}$ with Eq.~\ref{eq:calib}.

\section{Experimental results and analysis}
\label{sec:exp}

%In this section we validate our algorithm on three popular dataset.

In this section we thoroughly evaluate our algorithms and compare them with the-state-of-the-art methods on  public datasets.
In a second step, we discuss the impact of several important factors:
choice and number of reference images,
weighting scheme, architecture design, and model  complexity.

\vspace*{-10px}
\subsection{Datasets}
\vspace*{-1mm}
%We validated our algorithm on three public datasets. 

Since our method is designed for dealing with eye image alone,
without extra information from the face,
%or conjunction of two eyes. Thus
we considered the three following public eye-gaze datasets for validation.
%our algorithm. 

\mypartitle{\eyediap.}
It contains $94$ videos  from 16 subjects~\cite{funes2014eyediap}.
Videos belong to three categories: continuous screen (CS) target, discrete screen (DS) target or floating target (FT).
The CS videos were used in our experiments, which comprises static pose (SP) recordings (subjects approximately maintain
the same pose while looking at targets), and dynamic poses (MP, subjects perform additional important head movements while looking).
From this data, we cropped around $80K$ images of the left and right eyes and frontalized 
them according to~\cite{FunesMora2016}.
The labeled world gaze groundtruth was converted accordingly in the Headpose Coordinate System (HCS). 

\mypartitle{MPIIGaze.}
This dataset~\cite{zhang2015appearance} contains  $1500$ left and right eye images of 15 subjects, which were recorded
under various conditions in  head pose or  illuminations and contains people with glasses.
The provided images are gray scale and approximately of size $36\times60$ pixels, and are already frontalized relying on the head pose yaw and pitch. The provide gaze is labeled in Headpose Coordinate System (HCS). 
Note that although in~\cite{zhang2015appearance} the head pose was used as input for gaze prediction,
this did not improve our results in experiments so it was not used for the experiments reported below.

\mypartitle{UT-Multiview.}
This dataset~\cite{sugano2014learning} comprises  $23040$ (1280 real and 21760 synthesized) left and right eye samples for each of the 50 subjects
(15 female and 35 male).
It was collected under laboratory condition, with various head poses.
Eye images  are gray scale and of size  $36\times60$ pixels. 
They  are not frontalized but  accurate headpose and gaze in HCS are provided.
Thus, in experiments,  we concatenated the head pose in the network as described in~\cite{zhang2015appearance}.
More precisely, we concatenated the head pose $\headps(\Image)\in R^{1\times 2}$ of the input $I$ image
with the last fully-connected layers for the baseline CNN (Fig.~\ref{fig:gazecnn}), and did the same for the differential network,
i.e. we concatenated the two head pose $\headps(\Image)$ and $\headps(\Jmage)$ of the  input pair $(\Image,\Jmage)$
with the last fully-connected layer in Fig.~\ref{fig:siamese}.

\vspace*{-10px}
\subsection{Experimental protocol}
\vspace*{-1mm}

\mypartitle{Cross-Validation.}
For the \eyediap and MPIIGaze datasets, we applied a leave-one-subject-out protocol,
while due to its size,  we used a 3-fold cross-validation protocol for the UT-Multiview dataset.
Note that for this dataset, we train with real and synthesis data, but only test on real data.
Note that the protocols for MPIIGaze and UT-Multiview are the ones
from the original paper and followed by other researchers.

\mypartitle{Performance measure.}
Although nothing in the method prevents from using a single model for the left and right
eyes through eye image mirroring,
in experiments  we trained and tested models for the left and right eyes separately.
We noticed that there were some asymmetrical factors on the \eyediap dataset (see baseline results in Tab.~\ref{tab:result} for instance) 
probably caused by differences in the preprocessing (e.g. the face mesh may fit closer or further away on different parts of the face
depending on the viewpoint, which can affect the eye image normalization). 
Following the above protocols, the error was defined as the average of the average gaze angular error computed for each fold, according to~\cite{zhang2015appearance}.
%
%More precisely, if $\TestData$ denotes the test data (for a single subject) of a given fold, the trained model for that fold is evaluated by computing:
%\begin{align}
%    \mathcal{E}(\TestData) = \frac{1}{|\TestData|}\sum_{\Image\in \TestData} \arccos\Big( \GazeVecNorm\left(\GazePred(\Image)\right) \cdot \GazeVecNorm\left(\GazeGT(\Image)\right) \Big),
%\end{align}
%where $\GazeVecNorm(\theta_1, \theta_2)$ denotes the unitary 3D gaze vector associated with the gaze angles $(\theta_1, \theta_2)$.
%, which is defined as ${\bar{v}(\theta_1, \theta_2) = v(\theta_1, \theta_2)/\|v(\theta_1, \theta_2)\|}$ and 
%\begin{align}
%   v(\theta_1, \theta_2) = \left[ \sin{\theta_1}\cdot\cos{\theta_2}, \sin{\theta_2}, \cos{\theta_1}\cdot\cos{\theta_2} \right].
%\end{align}

\mypartitle{Selection of reference samples.}
For the linear adaptation and the differential NN methods,
%reference images are required to predict the  gaze of the given subject.
%In these cases,
unless stated otherwise, we randomly selected $9$ points as reference samples in the test set $\TestData$ for $200$ times,
and reported the average error computed for each random selection as defined above.

\mypartitle{Tested models.}
Several methods were tested for comparison.
\begin{itemize}
	\item {\bf \Baseline}: it corresponds to the generic model introduced in Section~2, and is our implementation of the neural network in~\cite{zhang2015appearance}, which achieves similar or better results than~\cite{zhang2015appearance}. Note that~\cite{zhang2017s} updates the result from~\cite{zhang2015appearance} using a much deeper VGG-16 network. For real-time purpose, we use shallow networks, so we use our generic model as baseline for a fair comparison.
	\item {\bf \SVRAd} is our implementation of the SVR adaptation method of~\cite{Krafka2016} built upon the \Baseline  model above. More precisely, following~\cite{Krafka2016}, the featuremap F2 (last layer before the output, see Fig.~\ref{fig:gazecnn}) is extracted as
          eye image features.
          A SVR model is trained using the reference image features and their gaze groundtruth.
%           and then applied on the test eye image features for prediction.
	\item {\bf \LinAd} corresponds to the \Baseline model followed by linear adaptation (Section 2.2).
	\item {\bf \DiffSiam}: our differential network, with the  default parameters introduced in the paper.
	\item {\bf \DiffFT}: differential network adapted via parameter finetuning using  reference samples (Section \ref{sec:lossadapt}).
	\item {\bf \DiffSiamout} differential network \DiffSiam  without the Gaussian kernel averaging (corresponds to \cite{liu2018differential}).
	\item {\bf \DiffVGG} differential network with a pre-trained VGG-16
          backbone (same learning parameters as \DiffSiam). 
\end{itemize}

%\begin{figure*}[!htp]
%	\centering
%	\subfigure[\eyediap]{\label{fig:result_a}
%		\includegraphics[width=0.32\textwidth]{images/eyediap_result.png}}
%	\subfigure[MPIIGaze]{\label{fig:resultb}
%		\includegraphics[width=0.32\textwidth]{images/mpii_result.png}}
%	\subfigure[UT-multiview]{\label{fig:resultc}
%		\includegraphics[width=0.32\textwidth]{images/utmv_result.png}}
%	\vspace*{-0mm}
%	\caption{Average angular error (degree) on three public datasets.
%		Note that the \Baseline  method does not require calibration data.
%	}
%	\vspace*{-3mm}
%	\label{fig:result}
%\end{figure*}

\vspace*{-10px}
\subsection{Experimental results}
\vspace*{-1mm}

\setlength{\tabcolsep}{0.25em} % for the horizontal padding
\begin{table}
	\centering
	\caption{Average angular error (degree) on three public datasets. `L, R, Avg' denote the left, right eyes and the average of them.
		Note that the \Baseline  method does not require calibration data.}
	\vspace{-5px}
	\label{tab:result}
	\begin{tabular}{|c|c|c|c|c|c|c|c|c|c|} \hline
		& \multicolumn{3}{c|}{\eyediap} & \multicolumn{3}{c|}{MPIIGaze} & \multicolumn{3}{c|}{UT-multiview}  \\\cline{2-10}
		& L & R & Avg & L & R & Avg & L & R & Avg \\\hline
		GazeNET~\cite{zhang2017s} & - & - & - & - & - & 5.5 & - & - & 4.4 \\\hline
		 \Baseline~\cite{zhang2015appearance} & 5.37 & 6.63 & 6.00 & 5.97 & 6.25 & 6.11 & 6.08 & 5.83 & 5.95 \\\hline\hline
	%	 & $\pm$2.7 & $\pm$2.8 & $\pm$2.8 & $\pm$1.2 & $\pm$0.9 & $\pm$1.1 & $\pm$1.6 & $\pm$1.6 & $\pm$1.6 \\ \hline 
		\SVRAd~\cite{Krafka2016} & 4.14 & 4.06 & 4.10 & 5.71 & 5.78 & 5.75 & 5.61 & 6.02 & 5.82 \\ \hline
	%	 & $\pm$1.3 & $\pm$1.2 & $\pm$1.2 & $\pm$1.3 & $\pm$1.2 & $\pm$1.3 & $\pm$1.2 & $\pm$0.8 & $\pm$1.0 \\\hline

		\LinAd & 3.88 & 3.81 & 3.84 & 5.68 & 5.66 & 5.67 & 4.57 & 4.56 & 4.56 \\  \hline
	%	& $\pm$1.6 & $\pm$1.6 & $\pm$1.6 & $\pm$1.4 & $\pm$1.3 & $\pm$1.4 & $\pm$0.1 & $\pm$0.1 & $\pm$0.1 \\\hline

		\DiffSiam & {3.23} & {3.23} & {3.23} & {4.69} & {4.62} & {4.64} & {4.17} & {4.08} & {4.13} \\ \hline
	%	& \textbf{$\pm$1.3} & \textbf{$\pm$1.2} & \textbf{$\pm$1.3} & \textbf{$\pm$1.2} & \textbf{$\pm$1.1} & \textbf{$\pm$1.1} & \textbf{$\pm$1.2} & \textbf{$\pm$0.9} & \textbf{$\pm$1.1}\\\hline\hline

  \DiffFT & \textbf{2.99} & \textbf{3.01} & \textbf{3.00} &  {4.61} & {4.56} & {4.59} &  \textbf{3.82} & {3.73} & \textbf{3.77}\\ \hline
	%	& \textbf{$\pm$0.9} & \textbf{$\pm$1.0} & \textbf{$\pm$1.0} & \textbf{$\pm$1.2} & \textbf{$\pm$1.1} & \textbf{$\pm$1.1} & \textbf{$\pm$1.0} & \textbf{$\pm$0.8} & \textbf{$\pm$0.9}\\\hline

	  \DiffSiamout & {3.37} & {3.35} & {3.36} & {4.73} & {4.61} & {4.67} & {4.41} & {4.24} & {4.33} \\ \hline

  \DiffVGG & {3.19} & {3.06} & {3.12} &  \textbf{3.88} & \textbf{3.73} & \textbf{3.80} & {3.88}  & \textbf{3.68} & {3.78} \\ \hline
	%	& \textbf{$\pm$1.2} & \textbf{$\pm$1.1} & \textbf{$\pm$1.2} & \textbf{$\pm$0.9} & \textbf{$\pm$0.8} & \textbf{$\pm$0.9} & \textbf{$\pm$1.1} & \textbf{$\pm$1.1} & \textbf{$\pm$1.1}\\\hline

	\end{tabular}
\vspace{-15px}
\end{table}

The experimental results are presented in Table~\ref{tab:result}.
%, in which the left, mid and right plots
%are the results on \eyediap, MPIIGaze and UT-multiview datasets.
%In each sub-figure, the upper bars indicate the results for the left eye,
%and the bottom ones for the right eye.
%%
%The colors correspond to the different approaches:
%\Baseline (blue),
%\LinAd (orange),
%\SVRAd~\cite{Krafka2016} (red),
%and our \DiffSiam proposed method (green).

\mypartitle{Baseline model.}
First, let us note that  under the same protocol, our \Baseline model
works slightly better than~\cite{zhang2015appearance}, which reported an error of $6.3\degree$ on MPIIGaze,
and of $5.9\degree$ on UT-Multiview.
%, compared to $6.11\degree$ and $5.95\degree$ in our case.
%
This is probably due to our network architecture being slightly more complex, while still avoiding overfitting.

\mypartitle{Linear and SVR Adaptation.}
Results demonstrate that, as expected, calibration helps and that the  
linear adaption method can greatly improve the baseline results, with an error decrease of 
(for the left and right eyes):
$27.7\%$ and $43.3\%$ on \eyediap,
$24.7\%$ and $21.8\%$ on UT-Multiview,
and $4.9\%$ and $9.4\%$ on MPIIGaze.
The difference in gain is most probably due to the recording protocols.
While the \eyediap and UT-Multiview datasets were mainly recorded over the course of one session,
the MPIIGaze dataset was collected in the wild, over a much longer period of time,
and with much more lighting variability (but less head pose variability).
This can be observed in Fig.~\ref{fig:bias} showing typical scattering plots of the \eyediap and MPIIGaze datasets.
%
%As can be seen,
The \eyediap plots follow a more straight and compact linear relationship than those on the
MPIIGaze dataset, reflecting the higher variability within the last dataset.
Seen differently, we can interpret the results as
having a session-based adaptation in the \eyediap and UT-Multiview cases,
whereas in MPIIGaze, the adaptation is more truly subject-based.

\begin{figure*}[tb]
	\centering   
	(a)\includegraphics[width=0.45\textwidth,height=30mm]{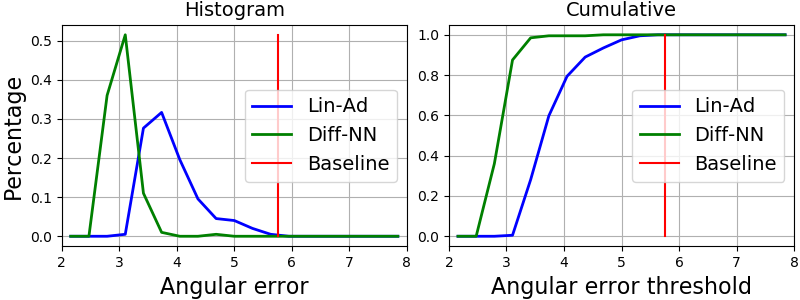}
        \hfill
	(b)\includegraphics[width=0.45\textwidth,height=30mm]{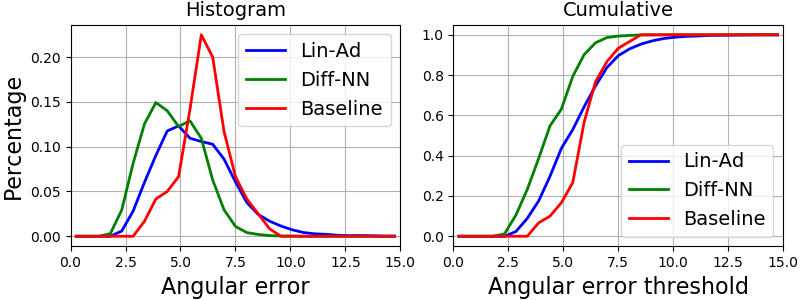}
	\vspace*{-5px}
	\caption{Histogram and cumulative histogram of angular errors (in degree) due to the random selection of the calibration images,
          for (a) a given user and  (b) all users of the MPIIGaze dataset,  and for different methods:
          \DiffSiam (green curve), \LinAd (blue curve), \Baseline (red; in (b) the average result per user is used for the plot).
	}
	\vspace*{-15px}
	\label{fig:randresult}
\end{figure*}

Results also show that the linear adaptation \LinAd method is working  better than the 
\SVRAd adaption approach~\cite{Krafka2016}, with an average gain of
% dataset    lin    svr         siam  (lin, %; svr %)
% eye  3.84  4.22   9%          3.36  (0.48   14.3     0.86,  25.6%; 
% mpii 5.67  5.74   1.4%        4.67  (1.00   21.4     1.07   23%;
% ut   4.56  5.81   21%         4.32  (0.24   5.5%;    1.49   34.5%
% AVGE                                        13.7%           27.7%
$6.3\%$, $1.4\%$ and $21.5\%$ on the \eyediap, MPIIGaze, and UT-Multiview datasets, respectively.
The main reason might be that in \SVRAd, 
the regression weights from the feature layer F2 are not exploited, in spite of their importance
regarding gaze prediction. 
In addition, finding an appropriate kernel in the $256$ dimensional space of F2 might not be so easy,
when using only  $9$ samples.
%points might not be sufficient for regression within such a space.

\mypartitle{Differential methods.}
Our approach \DiffSiam performs much better than the other two adaptation methods which, on average over the 3 datasets,
have an error $17.4\%$ (\LinAd) and $30.6\%$ (\SVRAd) higher than ours.
Interestingly, our method improves for all datasets and users on average\footnote{As for some bad calibration sample selections, results can be worse.} compared to \SVRAd,
and similarly compared to \LinAd with the exception of 2 users (out of 50) in UT-Multiview.
In particular, we can note that the gain is particularly important on the MPIIGaze dataset ($22.2\%$ compared to \LinAd),
demonstrating that our strategy of directly predicting the gaze differences from pairs of images
-hence allowing to implicitly match and compare these images-
using our differential network is more powerful, and more robust against eye appearance variations across 
time, places, or illumination, than adaptation methods relying on gaze predictions only (\LinAd), or on compact
eye image representations (\SVRAd).
To considering an even more realistic case, we randomly sampled the 9 reference samples from a single
day and tested on the other days. The performance only dropped from 4.64 to $4.83^{\circ}$ error, showing
the robustness of our method on this more 'subject-based' adaptation dataset.
On other more 'session-based' datasets, the linear adaptation method is already doing well,
so that the gain is lower (around $10\%$ on average).
Note that removing the weighting scheme of Eq.~\ref{eq:gazeest} (\DiffSiamout method) when combining per-reference gaze predictions   \cite{liu2018differential}
results in lower performance (around $0.2^{\circ}$), as further discussed in Sec. V.F.

%The siameses NN method, which predict the gaze difference between the given eye image and the reference images, outperforms other methods on all the three datasets. Especially on MPIIGaze dataset, it improves $20.8\%$ and $26.2\%$ from CNN model for left and right eyes respectively. We think this is because the siamese NN method is more robust on noise than linear adaptation. On \eyediap and UT-multiview datasets, the linear adaptation method has made a significant improvement, so we do not have a large margin on the result from siameses NN method to linear adaptation.
%
Further gains can be obtained with our differential method.
First, looking at the  \DiffFT results, we see that even with few reference samples (9) a systematic finetuning of the differential network 
can further improve the results: results of \DiffSiam are  7.5\%, 1\% and 9.5\% higher than those of  \DiffFT on \eyediap, MPIIGaze and UT-Multiview, respectively.
%
%
%% Besides, \DiffFT can improve the results slightly from \DiffSiam with few reference samples (9 samples used in this case). Especially on the EYEDIAP and the UTMultivew datasets, the angular errors are reduced by $7.1\%$ and $8.7\%$ respectively. The gain is not significant on the MPIIGaze dataset, we think it is caused by the larger noise in each subject on the MPIIGaze dataset.
%
Secondly and importantly, by simply using the deeper VGG-16 backbone to extract feature maps of the eye (instead of the C1-C3 CNN blocks, see Fig.~\ref{fig:siamese}),
we can reduce the errors by 4, 9 and 18\% on the \eyediap, UT-Multiview and MPIIGaze datasets, respectively.
This is obtained at the cost of a higher memory footprint and computational complexity.
%This
It makes the model competitive with respect to the state of the art: for instance the adaptation method in
\cite{Yu2019} reported an error of $4.2^{\circ}$ on the MPIIGaze dataset, compared to $3.8^{\circ}$ in our case.
%The difference might be due to a smaller image resolution input ($36\times 60$ in \cite{Yu2019}  vs $48\times 72$)
%or the use of more Max-pooling layers in our case.
%
The \DiffVGG  results  are similar to those of \DiffFT (except on MPIIGaze where it works much better).
It is left as future work to see whether finetuning would further improve the results.

\vspace*{-10px}
\subsection{Cross-dataset experiments.}

\vspace*{-1mm}

Such experiments are important and can be conducted to show a method generalization,
as long as the preprocessing and task formulation are equivalent
(e.g. addressing gaze estimation from face images, and using face datasets with the same
gaze definition).
Unfortunately, when working with existing cropped eye image datasets,
there are factors which can limit the validity of cross-dataset experiments,
as they clearly introduce systematic domain biases \cite{Yu:ECCVW:2018}.
Such factors include using different gaze coordinate systems and data preprocessing
methods, like  geometric normalization 
%or rectification
relying on different head pose estimators or cropping paradigms.

%this is the case of different image pre-processing steps, involving different head pose estimators, different eye image geometric rectifications which affect the geometric task we are addressing (3D gaze prediction from eye images taken from different viewpoints).
%regarding gaze estimation methods.
%However, we believe it is not a good setting in the context of gaze estimation from cropped eye images.
%Indeed, existing gaze datasets usually
%This introduces systematic biases between datasets regarding the gaze groundtruth 

Nevertheless, as our method relies on image pairs, one could hope
that it would be robust to these domain shifts.
To evaluate this, we trained methods using UT-Multiview and tested on MPIIGaze,
which share a similar normalization goal (compared to \eyediap) but not the same pre-processing.
 %
%  which share a similar normalization step 
 %(but not the same pre-processing).
%
Previous methods were reporting errors of $13.9^{\circ}$ \cite{zhang2015appearance} and $8.9^{\circ}$ \cite{xiong2019mixed}
without any reference samples. 
This paper baseline method achieves an angular error of $17.8^{\circ}$.
Errors of the \LinAd, \SVRAd and \DiffSiam methods are
respectively  9.2, 9.7 and 9.8  using 9 reference samples, 8.4, 8.1 and 8.4 with 50 samples.
 While all adaptation methods improve the baseline results significantly,
 their performance remain far from the within dataset  results (between 3.8 and 5.6).
 We believe this to be  due in great part to preprocessing
 discrepancies, and significant head and eye poses distributions difference between the two datasets.
 Unfortunately, our approach does not provide additional robustness against such geometric
 domain shifts.
 Handling them require methods of its own which can leverage
 more (labeled or no) target domain data.
 
%%  %
%%  %&we conducted cross-dataset experiments by
%% we trained methods using UT-Multiview and tested on MPIIGaze
%% which share a similar normalization goal (compared to \eyediap) but not the same pre-processing.
%%  %
%% %  which share a similar normalization step 
%%  %(but not the same pre-processing).
%%  %  
%%  The baseline method achieves an angular error of $17.8^{\circ}$, and adaptation methods achieve 9.2 (\LinAd), 9.7 (\SVRAd) and 9.8 (\DiffSiam).
%%  %
%%  While all adaptation methods improve the results from the baseline significantly,
%%  their performance remain very far from the within dataset  results (between 3.8 and 5.6),
%%  due in great part to dataset preprocessing discrepancies, as well as the fact that the
%%  distributions of head and eye poses differ significantly between the two datasets.
%%  %
%%  In practice some domain shift adaptation is needed to handle them.
 %se preprocessing discrepancies/biases. 

\vspace*{-10px}
\subsection{Impact of reference samples}
\label{sec:impactofref}
\vspace*{-1mm}

In this section, we discuss the impact of the selection and number of reference samples on performance. 
%We first analyse the variability associated to the choice of the related samples, and then study the number of samples required to achieve good results.

\mypartitle{Calibration data variability.}
The performance of the adaptation methods are computed as the average over 200 random selections
of  $9$ calibration samples.
Depending on the selection (samples might be noisy, or not distributed well on the gaze grid), results may differ.
The left plot of Fig.~\ref{fig:randresult}(a) shows the histogram of the angular error of \LinAd and \DiffSiam
for the different trials of one subject % (Fig.~\ref{fig:randresult}a))
and the right plot  shows the cumulative histogram (percentage of trials whose performance is below a threshold).
Fig.~\ref{fig:randresult}(b) does the same using the performance results of all users.

%
%illustrates the variabilities  of \LinAd and \DiffSiam for the different trials  of one subject % (Fig.~\ref{fig:randresult}a))
% and on average for all subjects.
%(Fig.~\ref{fig:randresult}b)).
%
%Fig.~\ref{fig:randresult_a} is the normalized histogram distributions of the angular error and the corresponding cumulative distributions respectively on one identity in \eyediap dataset. Fig.~\ref{fig:randresult_b} is about another identity in MPIIGaze dataset. In each figure, the red, blue and green lines denote the results of CNN, linear adaptation and siamese NN methods.
%

Fig.~\ref{fig:randresult}(a) shows an 
%typical
example where there is a relatively large bias for the given subject.
In that case, whatever the selection of the calibration samples,
the results of both  \LinAd and \DiffSiam are better than the baseline.
However, importantly,  our \DiffSiam approach is much less sensitive to the choice of calibration points than \LinAd,
as can be seen from the higher and concentrated peaks in the error distributions.
In other examples, the baseline is better (the red bar is within or more towards the peaks of the variability histograms),
but the  behaviors and relative placements of the \LinAd and \DiffSiam curves remain the same,
as shown by looking at the statistics over all users in Fig.~\ref{fig:randresult}(b).
%

%
%% The example on the right shows one of the few cases where the baseline is already good,
%% with little bias but nevertheless quite noisy samples.
%% %
%% In that case, there is only around  $60\%$ chances to obtain a better  result with the linear adaptation,
%% but still around $90\%$ chances with our approach.
%% %
%% Also, importantly,  our \DiffSiam approach is much less sensitive to the choice of calibration points than \LinAd,
%% as can be seen from the higher and concentrated peaks in the error distributions.

\begin{figure}[bt]
	\centering
	\includegraphics[width=0.8\linewidth]{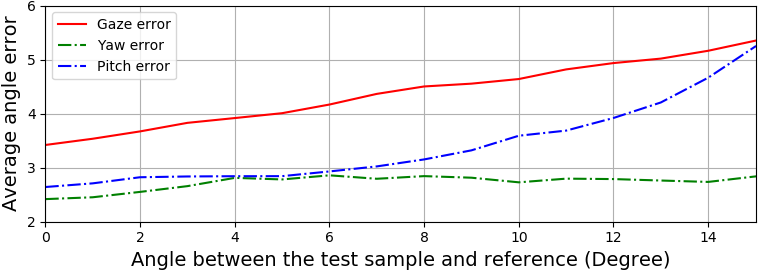}
	\vspace*{-5px}
	\caption{The average prediction error of the yaw in function of the absolute difference in yaw 
          between the reference and test samples (green curve), and similarly for the pitch and gaze.
	}
	\label{fig:error_dist}
	\vspace*{-15px}
\end{figure}

\mypartitle{Reference sample selection.}
%
%Fig.~\ref{fig:randresult} shows the histogram of average angular errors due to the random selection of the reference images, which indicates that some of the selection work better than others.
We can  analyze the impact of the reference samples % on performance
by measuring the error when using a single sample.
To do so, for each user, we randomly select $200$ times one sample as reference,
and then compute the errors for all test samples.
The resulting statistics for all users are shown in Fig.~\ref{fig:error_dist}, in which we plot
the average angular error of the yaw prediction (respectively pitch and gaze) in function of
the difference in yaw (respectively pitch and gaze) between the test and reference samples.

As expected, we observe that predictions are more accurate when  test samples are closer to the reference sample (red curve)
which justifies the use of the weighted sum in Eq.~\eqref{eq:gazeest}.
Interestingly, we also notice that the error profile of the yaw is relatively flat,
while that of the yaw is increasing.
%, especially after $10^{\circ}$.
% Looking more carefully at  results, we noticed the following.
%The yaw estimation  is particularly good, while the
%pitch one is sometimes more problematic.
%
%This is illustrated in Fig.~\ref{fig:error_dist}.
%\myremove{when considering the $s2$ reference sample,} the estimated pitch is not very good
%on average \myremove{(and even for $s0$ and $s1$, the errors are larger on  pitch than  yaw)}.
%
An explanation is that  when comparing two eye images,
aligning them laterally  relies on relatively stable structures (eye corners),
and the iris horizontal location (related to  yaw)  can be estimated
reliably from strong vertical edges.
However, visually, pitch estimation is much harder:
the vertical alignment relies on eyelid contours, which are moving structures (correlated with the pitch, but only partially),
%and thus less stable across eye images, and
% maesuring the change in vertical direction of the iris is more difficult,
and as the  iris top and bottom parts are often hidden,
the iris vertical position needs to be estimated from the shape of the iris vertical sides.

% suggesting that better weighting scheme should be exploited. This is left for future work.
	
	%Then we show the average angular error over the subject in Fig.~\ref{fig:error_dist2d} by using two reference samples. In this figure, the left and right plots indicate the yaw and pitch. The X-axis and Y-axis denote the two reference sample respectively. 
	
	%From the left plot for yaw error, the error is lower when the two reference samples are neither too far nor too close. However, it is different for pitch error in the right plot. We think it is because iris moves less when gaze pitch changes, which gives rise to a smaller changes of eyes. Overall, we think this is a good guideline to choose multiple reference samples.

%\begin{figure}[!bt]
%	\centering
%	\includegraphics[width=0.3\linewidth]{images/s0.png}
%	\includegraphics[width=0.3\linewidth]{images/s1.png}
%	\includegraphics[width=0.3\linewidth]{images/s2.png}
%	\includegraphics[width=0.45\textwidth,height=32mm]{images/siamese_error.png}
%		\vspace*{-5px}
%	        \caption{The top row shows three different reference image, $s0,s1, s2$.
%                  The bottom row plots the average angular prediction error (degree) of samples 
%	          in function of their absolute yaw (left plot) or pitch (right plot) when using a
%                  single reference sample $s0$ (red curve), $s1$ (green) and $s2$ (blue) for prediction.
%	}
%	\label{fig:error_dist}
%		\vspace*{-15px}
%\end{figure}

\begin{figure}[tb]
    \centering
    \includegraphics[width=0.4\textwidth]{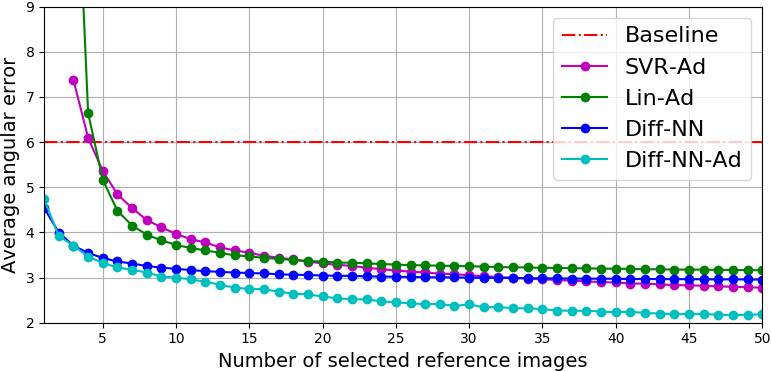}
	\vspace*{-5px}
    \caption{Comparison of average angular error in degree (average of the left right eyes) for different methods in function of the number of reference images on \eyediap dataset. Note that the \Baseline  method does not require calibration data.
     }
    \label{fig:mul_refer_result}
    	\vspace*{-15px}
\end{figure}

\mypartitle{Discussion about the number of references.}
Fig.~\ref{fig:mul_refer_result}  presents adaptation results on the \eyediap dataset using different number of reference images. % , which compares the \SVRAd, \LinAd, \DiffSiam with \Baseline.
When given few reference samples, the \SVRAd and \LinAd underperform the \Baseline, which is mainly due to the noise illustrated in Fig.~\ref{fig:bias}, which introduce a high variability (and error) in the fitting process, especially for \LinAd.
As the number of reference samples increases, the error of \LinAd decreases significantly because more accurate linear parameters can be obtained for adaptation.
The error of \SVRAd decreases more slowly  at the beginning, but catches up that of \LinAd when using more samples,
due to the inherent ability of  \SVRAd at  leveraging more reference samples.
%
% But with a large amount of reference samples, the performance of \SVRAd is almost equivalent to \LinAd. We infer that the \SVRAd inherently better leverage more reference samples.
%

The \DiffSiam outperforms the other methods for small numbers even when using only one reference samples.
This is not surprising because \DiffSiam does not learn any model or parameter from the reference samples,
but rather relies on richer information (the image context) to infer the difference rather than just the predicted gaze.
However, with more samples, the \SVRAd method works better, as the  Support Vector principle
might better handles larger amounts of data compared to our simpler weighting scheme,
and the \DiffSiam network prediction bias (and high variance) remains unchanged with more data.
Our network adaptation strategy \DiffFT does not have this limitation, and indeed
better leverage the availability of more reference samples.
While for small amount of reference samples the improvement over \DiffSiam is small (7.5\%),
with 25 samples, results of the \DiffSiam and \SVRAd methods are 23\% and 29\% higher than that of \DiffFT,
and with 50 samples they are 35\% and 27\% higher.
%, demonstrating its capacity at better leveraging reference samples.

% Another interesting point is that, given as many as $20$ reference samples, there is still a noticeable margin between the results of \DiffSiam and other methods.
%Although \DiffSiam works best, one drawback of \DiffSiam is that, as the number of reference samples increases, the time consuming also increases. While the time consuming will keep unchanged for \LinAd and \SVRAd in test stage. 

%% \mynew{
%%   With more reference samples, the \SVRAd will work better than \DiffSiam, because the network will fitted on this personal data.
%%   As a comparison, our \DiffSiam will converge to a limit.
%%   However, our \DiffFT outperforms all the rest by fine-tuning the network on this personal data if given more than $5$ reference samples and the margin between the results of \DiffFT and other methods turns significantly if given more than 25 reference samples.
%% }

\vspace*{-10px}
\subsection{Reference weighting scheme}
\label{sec:weight}
\vspace*{-1mm}

%
%The impacts of different reference image selections were presented in Sec.~\ref{sec:impactofref}.
%In this Section
We  analyse here the weighting scheme combining  the  predictions made from each reference image
(see Eq.~\eqref{eq:gazeest}) as well as the influence of the Gaussian kernel bandwidth $\sigma$.
Intuitively, when $\sigma$ is too small, mainly the closest reference sample will contribute to the prediction. Conversely, when $\sigma$ is too large, reference samples will have similar contributions. 
%In our experiment, $\sigma=0.1$ proves to be a good choice to all the test.

\begin{figure}[tb]
	\centering
	\includegraphics[width=0.4\textwidth]{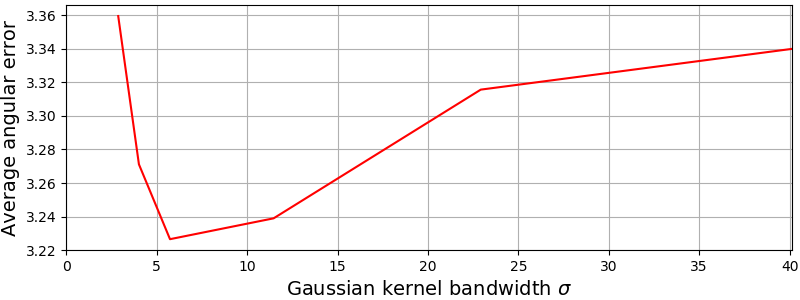}
		\vspace*{-5px}
	\caption{Average angular error (degree) on \eyediap dataset with different Gaussian kernel bandwidth $\sigma$ (in degree).
	}
	\label{fig:error_sigmas}
		\vspace*{-11px}
\end{figure}

Results are shown in Fig.~\ref{fig:error_sigmas}.
%, a Zero-mean Gaussian function $\mathcal{N}(0,\sigma)$ is used as the weight function. Empirically, we plot the average angular error on \eyediap dataset by using $9$ reference samples with different kernel $\sigma$ in Fig.~\ref{fig:error_sigmas}.
%
When $\sigma$ varies from $5.7$ to $22.8$ degrees, the error stays low, changing in a narrow range of $0.1$.
It indicates that the \DiffSiam is very robust w.r.t the weights, as also observed on MPIIGaze and UT-Multiview.
%datasets.
% Already mentionned in the method section
%In this paper, we choose $\sigma=0.1$, which gives the lowest error. 

%%===============================

\vspace*{-10px}
\subsection{System design}
\label{sec:archit}

\vspace*{-1mm}

Designing the network  architecture  can be motivated by several factors and principles.
The main idea behind  the differential network is that it can implicitly register and align images of the same eyes and
from there better compute the (differential) elements
(iris location, eye corners, eyelid closing) which really matters for gaze estimation than when abstracting these from a single image.
%
%A system should thus somehow explicitly or implicitly allow these type of processing.
%
%
We thus investigated different %schemes and different
levels at which to fuse the information coming from the two eye images.
%
%Early fusion can be achieved by concatenating the two eye images and using it as network input to the baseline architecture
%(See Fig.~\ref{fig:gazecnn}).
Early fusion is achieved by using concatenated eye images as  input to the network 
(See Fig.~\ref{fig:gazecnn}).
%It allows a direct comparison of the raw signals and fine geometric analysis,
It allows a direct comparison of the raw signals 
but suffers from  increase complexities  in (i) performing an implicit eye alignment
if eyes are too far apart in the input images; 
%have registration problem from the unprocessed images if they are not well aligned.
(ii) abstracting important eye structures for gaze difference prediction, 
because the information coming from the two eyes is mixed in the layers.
Also, the complexity increases, as all (input, reference) image pairs need to be fully processed.
At the other end, late fusion can be conducted by concatenating the  F2 feature maps of the two images.
The F2 eye representation might contain high level  eye representations tuned to the prediction of differential gaze,
but there might be a loss of localization information.
%

%% Based on this assumption, we propose the parallel architecture for differential network, in which the information of two images are fused to infer their differential gaze. As we know, it is different to fuse the information at different levels. 
%% %
%% On one hand, the early fusion, which concatenate the images direction, compare the raw signals and potentially allow for fine geometric analysis. 
%% %
%% But there are several drawbacks: a) it might have registration problem from the unprocessed images if they are not well aligned. b) Abstracting the important structures from the eyes in terms of prediction might be more difficult because the mixed information coming from the two eyes. c) The complexity increases, since all pairs of images (input, reference) need to be fully processed.

%% On the other hand, the late fusion, which concatenates featuremaps of F2 (dimension of 256) of the two images, can work with abstract representations tuned to the prediction of differential gaze.
%% %
%% But the drawback is that the representation is too abstract, giving rise to more difficulty for images registration and loss of localization informations.

Our proposed network lies in between.
It relies on  intermediate representations of each eye allowing in principle
the implicit registration of the two eyes from the processed images while extracting 
high level information relevant for differential gaze regression.
To verify the intuition behind  the proposed scheme, we compared the following  systems:
\begin{flushleft}
	\vspace*{-7px}
	\begin{compactenum}
		\item {\bf \archi{1} - early fusion}: concatenate the two images;
		\item {\bf \archi{2} - proposed }: concatenate the F1 feature maps;% of the two images;
		\item {\bf \archi{3} - late fusion}: concatenate the F2 feature map; %  of the two images;
		\item {\bf \archi{4} - siamese}: two parallel \Baseline networks with shared weights
					trained to only predict gaze differences.
		\item {\bf \archi{5} - multi-task with adaptation}: it corresponds to \archi{4} but trained to predict
					both the absolute gaze and the difference            (as in ~\cite{venturelli2017depth} for head pose).
					The trained  network is further adapted using the \LinAd scheme.
	\end{compactenum}
\vspace*{-3px}  
\end{flushleft}

\setlength{\tabcolsep}{0.3em} % for the horizontal padding
\begin{table}[bt]
	\centering
	\caption{Average angular error (degree) on \eyediap dataset for different systems (see text).
%        \archi{2} is the proposed  one.
%        For \archi{5}, we estimate the gaze directly as in~\cite{venturelli2017depth} and then adapt it with the \LinAd scheme.
        }
	\label{tab:archi_result}
        \vspace*{-1mm}
	\begin{tabular}{c|c|c|c|c}\hline
		\archi{1}-early & \archi{2}-proposed & \archi{3}-late & \archi{4}-siamese & \archi{5} \\\hline
		3.40 & 3.23 & 3.47 & 3.40 & 3.76 \\\hline
	\end{tabular}
    %   \vspace*{-3ex}
\end{table}

Results are shown in Table~\ref{tab:archi_result}, and show that our architecture achieves the best results.
Note that \archi{5} (approach of~\cite{venturelli2017depth} followed by \LinAd) is worse than all other systems,
demonstrating  the advantage of predicting differential gaze over absolute gaze. 
The results of the \archi{1,3,4} are close but still outperformed by our system  \archi{2}, showing that intermediate fusion
is better than early or late fusion.
We believe this is due to the ability of the CNN network layers to do some filtering and alignment of the two images,
while the fully-connected layers combine this information to infer gaze differences.

\vspace*{-10px}
\subsection{Algorithm complexity}

\vspace*{-1mm}

The \DiffSiam adaptation method does not have the same complexity as the others.
Compared to the CNN \Baseline, the linear adaptation only requires the computation of  Eq.\eqref{eq:linear},
which has negligible computational cost.
Our \DiffSiam approach, however, requires to predict the gaze differences between the test sample and \NRefData
reference images.
Fortunately,  its  complexity is not  \NRefData times that of the \Baseline 
thanks to our differential architecture (see Fig.~\ref{fig:siamese}).
%, the extra-computation is not as high.
%
%Indeed, first we can pre-compute and save the feature maps at the last convolutional neural layer of all the reference images,
% so that the computation of one gaze difference requires mainly the forward pass of one image.
%
% Secondly, the feature maps of the test image also need to be computed only once, which can be achieved by
% stacking the feature maps of the reference images in a mini-batch, and compute all gaze differences in parallel.
%
Indeed,  we can pre-compute and save the feature maps at the last convolutional neural layer of all the reference images.
Thus, the complexity reduces to the computation of the feature maps of the input image
and of \NRefData gaze differences from the feature maps, which can be done in parallel
within a mini-batch.

Table~\ref{tab:runtime} compares the running time (in ms) for the \Baseline and the
different \DiffSiam options (and $\NRefData=9$).
They have been obtained by computing the average run-time of processing  $5000$ images.
The CPU is an Intel(R) Core(TM) i7-5930K with 6 kernels and 3.50GHz per kernel.
The GPU is an Nvidia Tesla K40.
The program is written in Python and Pytorch.
Note that as the Pytorch library will call multiple kernels for computation, the CPU-based run-time is also short.
From this Table, we can see that our \DiffSiam  method and architecture
has a computational complexity close to the \Baseline.

\setlength{\tabcolsep}{0.2em}
\begin{table}[tb]
  \centering
  \vspace{-1mm}
    \caption{Run-times (in ms) between the \Baseline and our \DiffSiam method,
             using mini-batch (\DiffSiam$\!^{*}$) computation or not.}
    \label{tab:runtime}
  \vspace{-1mm}
    \begin{tabular}{|c|c|c|c|c|c|c|} \hline
         & \multicolumn{3}{c|}{CPU} &  \multicolumn{3}{c|}{GPU} \\ \hline
         & \Baseline  & \DiffSiam  & \DiffSiam$^{*}$  & \Baseline  & \DiffSiam & \DiffSiam$^{*}$ \\\hline
    Run-time & 2.5 & 7.6 & 3.5 & 1.4 & 4.0 & 1.5 \\ \hline
    \end{tabular}
  \vspace{-2mm}
\end{table}

%% \mynew{
%% Moreover,	we indeed can use a pre-trained network, such as the VGG-16 as backbone~\cite{Zhang2017a}, to improve our results. More precisely, $18\%$ improvement on MPIIGaze and $8.4\%$ on UTMultivew in Tab.~\ref{tab:result}. The reasons we don't apply large backbones are 1) we emphesize the differential structure in our paper, instead of the backbones; 2) various experiments have been tested and validated on this structure, and it will be super time/energy comsuming if using a large network such as VGG-16 as backbone; 3) results using VGG-16 in Tab.~\ref{tab:result} are persuaded.
%% }

\vspace*{-2mm}

\section{Conclusion}
\label{sec:con}

\vspace*{-1mm}

This paper aims to improve appearance-based gaze estimation using subject specific models built from few
calibration images.
Our main contribution is to  propose a differential NN for predicting  gaze differences instead of gaze directions
to alleviate the impact of annoyance factors like illumination, cropping variability, variabilities in eye shapes.
Experimental results on three public and commonly used datasets prove the efficacy of the proposed methods.
More precisely, while
%at very little extra computation cost
standard linear adaptation method can already boost the results on single session like situations,
the differential NN method produces even more robust and stable results across different sessions of the same user,  but costs some
more run-time compared to a baseline CNN.
Further fine-tuning  of the network using the reference samples
provide as well as very good mean to leverage larger amounts
of calibration samples.

% % use section* for acknowledgment
% \ifCLASSOPTIONcompsoc
%   % The Computer Society usually uses the plural form
%   \section*{Acknowledgments}
% \else
%   % regular IEEE prefers the singular form
% \fi

% This work was partly funded by the UBIMPRESSED project of the Sinergia interdisciplinary program of the Swiss National Science Foundation (SNSF), and by the the European Union’s Horizon 2020 research and innovation programme under grant agreement no. 688147 (MuMMER, mummer-project.eu).

\ifCLASSOPTIONcaptionsoff
  \newpage
\fi



%\bibliographystyle{IEEEtran}

\bibliographystyle{IEEEtran}
%\newpage
\bibliography{egbib}

\end{document}